\documentclass{article} 
\usepackage{iclr2027_conference,times}

\usepackage{amsmath,amsfonts,bm}

\def\eqref#1{equation~\ref{#1}}

\def\1{\bm{1}}

\DeclareMathAlphabet{\mathsfit}{\encodingdefault}{\sfdefault}{m}{sl}
\SetMathAlphabet{\mathsfit}{bold}{\encodingdefault}{\sfdefault}{bx}{n}

\usepackage[hidelinks]{hyperref}
\usepackage{url}
\usepackage{amsmath}
\usepackage{booktabs}
\usepackage{tabularx}
\usepackage{array}
\usepackage{graphicx}

\newcolumntype{Y}{>{\raggedright\arraybackslash}X}

\title{ERIA: An Instruction-Optimized Agent for Clinical EEG Reporting}

\author{
Ahmed Sameh$^{1}$,
Gayal Kuruppu$^{1}$,
Neeraj Wagh$^{2}$,
Bridger Brundy$^{3}$,
Yogatheesan Varatharajah$^{1}$ \\
$^{1}$Department of Computer Science \& Engineering, University of Minnesota Twin Cities \\
$^{2}$Department of Bioengineering, University of Illinois Urbana-Champaign \\
$^{3}$Department of Genetics, Cell Biology \& Development, University of Minnesota Twin Cities
}

\iclrfinalcopy
\begin{document}

\maketitle
\lhead{Preprint}

\begin{abstract}
Clinical EEG interpretation is substantially more complex than classifying short signal segments; it must assess technical quality, characterize background and behavioral states across long recordings, identify interictal and ictal abnormalities, reconcile uncertain evidence, and synthesize these findings into a clinically meaningful report. We introduce ERIA (EEG Reporting and Instruction-Optimized Agent), a locally deployable EEG interpretation framework that enables a single 27B language-model controller to analyze whole-recording EEGs by orchestrating specialized parametric and signal-processing tools. ERIA combines specialized tools for detecting background abnormality, sleep--wake states, seizure-like activity, and interictal epileptiform discharges with deterministic spectral, amplitude, symmetry, montage, and signal-quality analyses, covering multiple clinically relevant domains reflected in EEG reporting guidelines. Rather than relying on EEG-specific controller fine-tuning, a proprietary frontier model, or multi-agent orchestration, ERIA keeps the controller weights, specialist tools, and safety constraints fixed and optimizes only its natural-language analysis policy with GEPA (Genetic-Pareto), a reflective prompt optimizer, using feedback from expert-written EEG reports. Compared with the raw policy, GEPA substantially improves physician-aligned whole-recording report quality, while the resulting policy transfers to external EEG benchmarks. We evaluate performance on \emph{CerebraGloss-Bench} short-segment clinical interpretation and \emph{BrainBench} Foundational Analysis, where ERIA performs competitively with the models evaluated in the original benchmarks. These results show that carefully selected specialist tools and language-level policy optimization can extend the clinical and general EEG capabilities of a single locally executable model while preserving local inference, traceable tool use, and a comparatively simple deployment architecture.
\end{abstract}

\section{Introduction}
Expert visual review of patient data is still the standard practice for many clinical data modalities. This burden not only contributes to physician burnout but also introduces reviewer bias and errors in clinical decisions \citep{agarwal1998automatic}. This is especially true in neurology, where experts spend a significant amount of time visually reviewing and interpreting lengthy multi-channel time series of brain electrical activity, known as electroencephalography (EEG). EEG is a fundamental technology in neurological care, particularly in epilepsy, where it is one of the earliest tests performed due to its convenience and cost-effectiveness. However, expert EEG review is time-consuming, laborious, and error prone. As a result, there is a significant opportunity to augment the clinical EEG review process with machine learning (ML) \citep{tveit2023automated}. 

However, there are several challenges associated with developing a clinical-grade ML-based EEG review system. Clinical EEG interpretation is fundamentally a whole-recording reporting problem and must integrate technical quality, patient behavioral states, background activity, focal and generalized abnormalities, and an overall clinical interpretation. Established EEG reporting guidance from the American Clinical Neurophysiology Society (ACNS) reflects this breadth and emphasizes both systematic description of the recording and synthesis of the findings into a clinically meaningful interpretation \citep{tatum2016guidelines,kane2017glossary}. Accurate event detection alone is therefore insufficient; evidence must be gathered across the recording and integrated into a coherent report. On the other hand, a practical clinical system must also operate under deployment constraints. Local inference can be especially attractive for sensitive clinical data because it allows institutions to keep patient information within their own infrastructure and reduces reliance on external model providers \citep{griot2025implementation}. This creates a practical tension: clinical EEG systems should support broad, evidence-based interpretation without requiring excessively large or complex deployment stacks.

Existing work has approached this problem from two complementary directions. The first develops EEG-language systems that produce clinical descriptions, reports, or interpretations directly from EEG signals or learned EEG representations. This direction includes report-generation systems such as EEGtoText \citep{biswal2019eegtotext} and CELM \citep{pradeepkumar2026celm}, as well as vision-language interpretation models such as CerebraGloss \citep{gu2026cerebragloss}. Collectively, these approaches demonstrate increasingly capable mappings between EEG and clinical language, including long-recording report generation and fine-grained interpretation. These approaches, however, remain encoded within a fixed learned model or inference pipeline rather than an adaptive process in which a controller can actively select, combine, and revisit complementary specialist analyses. A second direction treats EEG and broader neuroscience analysis as an agentic process. Systems such as EEGAgent \citep{zhao2026eegagent}, BrainAgent \citep{zhou2026brainagent}, and NS-Copilot \citep{liu2026nscopilot} use language-model reasoning to coordinate analytical tools, pretrained models, or executable workflows, while BrainBench extends evaluation toward comprehensive instruction-conditioned EEG analysis \citep{zhou2026brainbench}. These approaches make evidence acquisition and multi-step analysis more flexible, but current systems typically rely on at least one of very large controllers, hosted frontier models, multi-agent coordination, or computationally heavier closed-loop execution. Moreover, whole-recording clinical report quality has not been rigorously evaluated against paired physician-authored references in this agentic setting.

Here we introduce \textbf{ERIA}, an \textbf{E}EG \textbf{R}eporting and \textbf{I}nstruction-Optimized \textbf{A}gent, to bridge these two directions. ERIA retains the explicit evidence gathering of a tool-using agent while using a single 27B controller designed for local execution. Rather than mapping EEG representations directly into report text, the controller reasons over outputs from specialist EEG analyses and determines how those findings should be combined into a clinical interpretation. Its toolset is selected to cover the principal information categories required for clinical EEG reporting according to the ACNS guidelines \citep{tatum2016guidelines,kane2017glossary}. We improve ERIA without EEG-specific fine-tuning of the controller or additional language-model agents. The controller weights, specialist tools, and safety constraints remain fixed; only the natural-language policy governing analysis and report synthesis is optimized. We use GEPA (Genetic-Pareto) to improve this policy from execution trajectories and textual feedback \citep{agrawal2026gepa}. Physician-authored reports provide the feedback signal, making whole-recording clinical interpretation the direct optimization target.

Experiments show that GEPA substantially improves physician-aligned whole-recording report quality over ERIA's raw policy and that the improvement transfers beyond the optimization task. The optimized policy achieves competitive performance on CerebraGloss-Bench for short-segment clinical interpretation and on BrainBench Foundational Analysis for broader EEG analysis \citep{gu2026cerebragloss,zhou2026brainbench}. These results suggest that improving how a grounded EEG agent gathers and integrates evidence can benefit both clinical reporting and broader EEG-analysis capability. Our contributions are threefold: (1) we introduce a grounded single-controller architecture for local EEG analysis; (2) we develop a physician-report-driven approach to optimizing tool-use policy without EEG-specific controller fine-tuning; and (3) we show that the resulting policy improves both clinical EEG reporting and transfer to independent EEG benchmarks.

\section{Related Work}

\paragraph{EEG-to-Language and Clinical Report Generation.}

EEG-to-language systems aim to translate neural recordings into clinically meaningful text. EEGtoText predicts a predefined set of EEG phenotypes and uses them to generate structured report sections, with attention providing localization of supporting signal regions \citep{biswal2019eegtotext}. CELM extends report generation to long-duration recordings by encoding EEG into latent representations that are aligned with a local language model and decoded into structured clinical reports \citep{pradeepkumar2026celm}. Although effective for long-context reporting, CELM follows a learned latent EEG-to-language pathway rather than exposing independently interpretable specialist analyses that can be selectively queried or revisited during inference. CerebraGloss takes a different approach, fine-tuning a vision-language model on rendered EEG waveforms and clinical instruction--response data for fine-grained interpretation of short EEG segments \citep{gu2026cerebragloss}. These approaches demonstrate strong EEG-to-language capabilities, but their analytical pathways remain fixed after training rather than being dynamically assembled from complementary specialist analyses.

\paragraph{Agentic EEG and Neuroscience Analysis.}

Agentic systems instead allow a language model to control the analytical process. EEGAgent is the closest architectural predecessor to ERIA, but relies on the substantially larger Qwen3-235B controller to dynamically select parametric and signal-processing tools, inspect their outputs, retrieve domain knowledge, and generate structured EEG reports \citep{zhao2026eegagent}. Its report-generation experiment is presented as a feasibility study because matched EEG--report pairs were unavailable; consequently, its analysis policy was neither optimized using clinical-report feedback nor quantitatively evaluated against paired references.

Other work increases analytical flexibility through multi-agent orchestration. BrainAgent uses a hierarchical supervisor-and-specialist architecture for broader brain-signal analysis and is evaluated with large and proprietary language-model backbones, including Qwen-Max and GPT-4o \citep{zhou2026brainagent}. NS-Copilot similarly coordinates multiple agents for model selection, code generation, execution diagnosis, and iterative replanning; all four of its agents are powered by GPT-5.2, with GPT-5.2-Codex additionally used to establish its execution threshold \citep{liu2026nscopilot}. BrainBench further broadens evaluation toward instruction-conditioned EEG analysis across multiple domains and provides a useful testbed for agentic EEG systems \citep{zhou2026brainbench}. In contrast, ERIA focuses on a single locally executable 27B controller and asks whether its clinical and general EEG capabilities can be improved through better tool-use policy rather than additional agents or scale.

\paragraph{Language-Level Optimization of Agent Policies.}

Language-model behavior can also be improved without modifying model weights. Reflexion uses linguistic feedback from previous attempts to guide subsequent decisions \citep{shinn2023reflexion}, while Self-Refine iteratively generates feedback and revises outputs without additional training \citep{madaan2023selfrefine}. GEPA extends this principle to optimization of language-model programs by reflecting over execution trajectories and textual feedback to propose improved natural-language instructions \citep{agrawal2026gepa}. ERIA applies this idea to grounded clinical EEG analysis by optimizing the policy governing evidence acquisition, verification, and synthesis, with expert-written EEG reports providing the feedback signal.

\section{ERIA}
\label{sec:eria}

\begin{figure}[!h]
    \centering
    \includegraphics[width=0.8\linewidth]{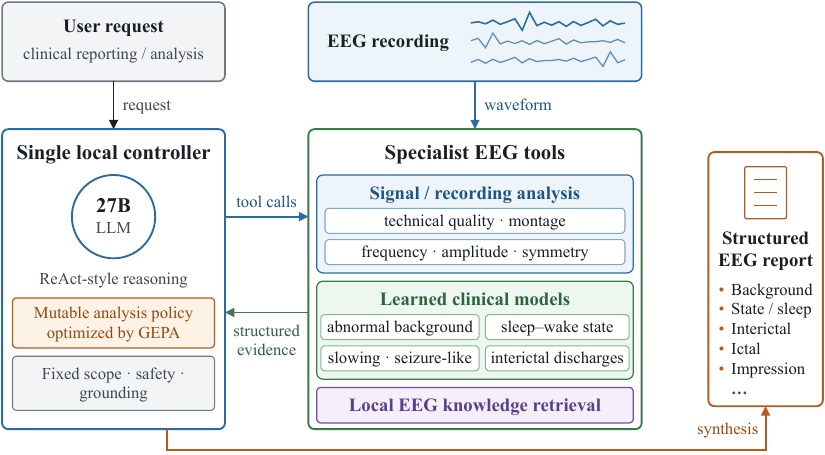}
    \caption{
    ERIA's inference workflow. A single local language-model controller selects specialist EEG analyses and reasons over their structured outputs rather than consuming waveform samples directly. The resulting evidence is iteratively gathered and synthesized into a structured EEG report.
    }
    \label{fig:eria_workflow}
\end{figure}

ERIA analyzes EEG through a single language-model controller that orchestrates specialist signal-analysis and clinical tools. The controller does not consume waveform samples directly; instead, it receives the user request, selects tools, observes their structured outputs, and gathers additional evidence as needed. Figure~\ref{fig:eria_workflow} summarizes the agent workflow.

\paragraph{Single-Controller Agent.}

ERIA uses a locally served Qwen3.8-27B controller \citep{qwen2026qwen38} in a ReAct-style reasoning-and-action loop \citep{yao2023react}. At each iteration, the controller can issue one or more tool calls with temporal, spatial, or analysis-specific arguments. Tool outputs are returned to the controller and incorporated into subsequent reasoning. This cycle continues until the controller determines that sufficient evidence has been gathered to answer the request. For clinical reporting, the collected evidence is synthesized into a structured EEG report. 

The system prompt is separated into a fixed component and a mutable analysis policy. The fixed component specifies the system scope, safety constraints, and evidence-grounding rules, which require learned-model outputs to be treated as supporting evidence rather than direct diagnoses. The mutable policy instead governs how the controller explores the recording, selects and revisits analyses, reconciles complementary evidence, and constructs the final response. GEPA modifies only this policy. For report-generation requests, ERIA follows the structured format used by the reference EEG reports in our paired clinical dataset, covering technical information, background activity, activation procedures, sleep, interictal activity, ictal activity, impression, and clinical correlation.

\paragraph{Guideline-Informed Specialist Toolset.}

The tool library is organized around evidence categories emphasized in ACNS Guideline~7 for routine EEG reporting \citep{tatum2016guidelines}. ERIA operationalizes these categories through complementary deterministic analyses, learned clinical models, and local knowledge retrieval. Table~\ref{tab:eria_tools} summarizes the core EEG-analysis tools. For several learned tools, we did not identify a released model that simultaneously matched the required output space and provided suitable performance for the intended role. We therefore trained task-specific models or prediction heads on external public datasets where necessary, while reusing released models when an appropriate model was available.

\begin{table}[t]
\small
\caption{Core ERIA EEG-analysis tools. ``Ours'' indicates that the task-specific
model or prediction head was trained by us. Detailed training procedures are
provided in the appendix.}
\label{tab:eria_tools}
\centering
\setlength{\tabcolsep}{4pt}
\renewcommand{\arraystretch}{1.08}
\begin{tabularx}{\linewidth}{
@{}
>{\raggedright\arraybackslash}p{0.22\linewidth}
>{\raggedright\arraybackslash}p{0.36\linewidth}
>{\raggedright\arraybackslash}p{0.16\linewidth}
Y
@{}}
\toprule
\textbf{Tool} & \textbf{Method / source} & \textbf{Scale} & \textbf{Returned evidence} \\
\midrule
\multicolumn{4}{@{}l}{\textit{Deterministic / inspection tools}} \\
\midrule
\path{recording_summary}
& Deterministic metadata inspection
& recording; all ch.
& metadata, duration, sampling rate, channels \\

\path{assess_signal_quality}
& Deterministic quality analysis
& interval; selected/all ch.
& invalid, flat, or clipped channels; quality summary \\

\path{inspect_preprocessing}
& Deterministic preprocessing inspection
& recording; all ch.
& preprocessing and transformed-signal statistics \\

\path{inspect_montage}
& Deterministic montage derivation
& recording; 22-ch TCP
& montage feasibility; derived channels \\

\path{amplitude_statistics}
& Deterministic amplitude features
& 10--60\,s; 1 ch.
& mean abs.\ amplitude, RMS, peak-to-peak, extrema \\

\path{spectral_analysis}
& Welch PSD
& 10--60\,s; 1 ch.
& band power; dominant-rhythm summaries \\

\path{hemispheric_symmetry}
& Left--right Pearson correlation
& 10--60\,s; homologous pairs
& interhemispheric correlation; significance \\

\midrule
\multicolumn{4}{@{}l}{\textit{Learned clinical tools}} \\
\midrule
\path{background_abnormality}
& ShallowFBCSPNet (ours); TUAB~\citep{lopez2015abnormal,obeid2016tuh,schirrmeister2017deep}
& 30\,s; 19 ref.\ ch.
& normal/abnormal probabilities \\

\path{coarse_activity_screen}
& SPaRCNet + LR head (ours); TUSL/TUSZ/TUAR~\citep{jing2023sparcnet,vonweltin2017slowing,shah2018tusz,hamid2020tuar}
& 10\,s; 16 bipolar ch.
& background/slowing/seizure-like probabilities; candidate intervals \\

\path{fine_event_classifier}
& XGBoost on handcrafted features (ours); TUSZ/TUAR~\citep{chen2016xgboost,shah2018tusz,hamid2020tuar}
& 2.5\,s; 22-ch TCP
& background/artifact/seizure-like probabilities; candidate intervals \\

\path{sleep_wake_staging}
& SleepStagerChambon2018 (ours); Sleep-EDF + EEGMMIDB~\citep{chambon2018sleep,kemp2000sleepedf,schalk2009eegmmidb}
& 30\,s; Fpz--Cz, Pz--Oz
& wake/drowsiness/sleep probabilities \\

\path{ied_candidate_screen}
& SpikeNet2 (released); no ERIA fine-tuning~\citep{li2025spikenet2}
& 1\,s, 0.25\,s stride; 19$\rightarrow$37 ch.
& IED probabilities; candidate intervals \\

\midrule
\multicolumn{4}{@{}l}{\textit{Knowledge retrieval}} \\
\midrule
\path{retrieve_eeg_knowledge}
& Local sentence-embedding retrieval over a curated EEG knowledge base
& query-level
& relevant EEG reporting and interpretation passages \\
\bottomrule
\end{tabularx}
\end{table}

Following the coarse-to-fine temporal analysis strategy used in EEGAgent \citep{zhao2026eegagent}, ERIA combines tools operating at different temporal resolutions. The 10-s coarse screen can survey longer intervals for slowing or seizure-like activity, after which the controller can inspect candidate regions using the 2.5-s event classifier. Sleep and background are assessed over 30-s epochs, whereas the IED screen uses densely overlapping 1-s windows. For longer scans, adjacent positive windows are merged into candidate intervals that the controller can inspect at finer resolution.

Technical reliability is also exposed to the controller rather than hidden inside the tools. Each analysis applies its own training-matched preprocessing and signal-quality checks and returns a quality note or warning when the analyzed signal required an intervention or may limit interpretation. For example, if a tool imputes a small number of missing samples before inference, that intervention is included in the returned quality note. This allows ERIA to account for technical uncertainty when interpreting tool outputs.

\paragraph{Physician-Aligned Clinical Evaluator.}

Optimizing an agent for clinical reporting requires a report-level objective rather than the score of any individual detector. We therefore use a separate LLM-based clinical evaluator that compares an ERIA-generated report with its clinical reference. The evaluator receives only the generated report and reference report. It scores seven clinical domains using $\{0,1,2,\mathrm{NA}\}$, where 0 denotes incorrect, 1 partially correct, 2 clinically equivalent, and NA indicates that the domain cannot be assessed from the physician reference report. The evaluated domains are background, ictal activity, interictal epileptiform activity, overall classification, state/sleep, focal or regional abnormality, and impression. Activation procedures are excluded from evaluation because their performance and timing depend on the acquisition protocol and are not available to ERIA from the EEG waveform alone. We use fixed weights of $0.25$, $0.20$, $0.15$, $0.10$, $0.10$, $0.10$, and $0.10$, respectively. The weights are set a priori to place greater emphasis on background and ictal/interictal findings while retaining coverage of the remaining reporting domains, consistent with the central role of these elements in established EEG reporting frameworks \citep{tatum2016guidelines,beniczky2013score}.

For aggregation, evaluator scores are normalized to $[0,1]$, with NA domains excluded. The aggregate score for case $i$ is $S_i = \sum_d w_d\,s_{i,d} \,/\, \sum_d w_d$, where $d$ ranges over $\mathcal{D}_i$, $w_d$ is the fixed weight assigned to domain $d$, $s_{i,d}\in\{0,0.5,1\}$ is the normalized domain score, and $\mathcal{D}_i$ is the set of domains that are applicable and assessable from the physician report. Renormalizing over $\mathcal{D}_i$ prevents unavailable reference information from lowering the score. In addition to the numerical domain scores, the evaluator produces domain-specific feedback and an overall textual critique.

\begin{figure}[t]
    \centering
    \includegraphics[width=0.8\linewidth]{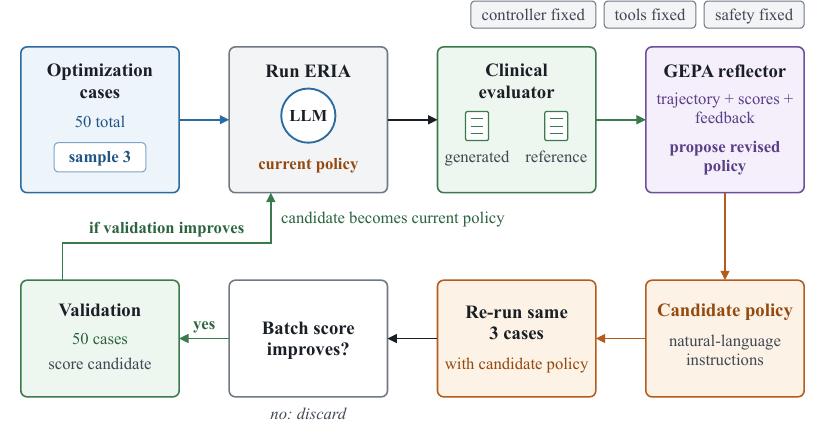}
    \caption{
    GEPA optimization of ERIA's natural-language analysis policy. The reflector uses execution trajectories and clinical-evaluator feedback to propose policy revisions. Candidate policies are first tested on the sampled optimization cases and, when improved, evaluated on the validation partition for selection. The controller, specialist tools, and safety constraints remain fixed, and the protected test set is inaccessible throughout optimization.
    }
    \label{fig:gepa_workflow}
\end{figure}

\paragraph{GEPA Policy Optimization.}
ERIA uses GEPA to improve the mutable analysis policy \citep{agrawal2026gepa}. We refer to ERIA's initial mutable analysis policy before GEPA optimization as the \emph{raw policy}. Figure~\ref{fig:gepa_workflow} summarizes this process. For each candidate policy, the frozen ERIA agent executes the EEG-analysis task and produces both a complete execution trajectory and a final report. The clinical evaluator independently compares the report against the physician reference and returns the domain scores and textual feedback described above.

The LLM-based GEPA reflector receives the complete execution trajectory, generated and reference reports, evaluator domain scores and feedback, and aggregate score. Fixed safety instructions and tool descriptions are provided as read-only context, while GEPA modifies only the mutable analysis policy. Candidate policies are evaluated on sampled optimization cases and, when improved, on the validation partition for selection. The protected test partition remains inaccessible throughout optimization, after which the selected policy is frozen for subsequent clinical and external evaluations.

\section{Experimental Setup}
\label{sec:experiments}

\paragraph{Internal Evaluation.} We first evaluate ERIA using a private dataset of 200 routine clinical EEG studies and matched clinically-generated reports. These EEG reports were authored by board-certified epileptologists according to the ACNS guidelines. Following the optimization protocol described above, the dataset is split at the recording level into 50 optimization cases, 50 validation cases, and 100 protected test cases. Note that this dataset was not used to develop any of the tools that the controller uses to generate reports.

For all experiments, ERIA utilized the Qwen3.8-27B controller served locally through Ollama with temperature 0, a 65,536-token context window, a maximum output length of 8,192 tokens, and at most 60 ReAct iterations per request. To compare ERIA's reports with epileptologist reports, we utilized a physician-aligned evaluator, which was instantiated with GPT-OSS-120b using high reasoning effort and temperature 0. We selected this model because it combines structured-output capabilities with strong performance on health-oriented evaluation, including HealthBench \citep{openai2025gptoss}. Its suitability for clinical-text review is further supported by a study evaluating GPT-OSS-120b as a second reviewer of real-world internal-medicine reports \citep{skrabic2026secondreviewer}.

\underline{Optimization Setup}: GEPA reflection uses \texttt{Qwen3.5-122B-A10B-NVFP4}. Because reflection operates over long execution contexts and requires synthesizing complex intermediate evidence, we selected Qwen3.5-122B-A10B for its strong reasoning, instruction-following, and long-context capabilities \citep{qwen2026qwen35}. The NVFP4 checkpoint enables efficient local execution of the reflector. GEPA is initialized from ERIA's raw mutable analysis policy and uses a budget of 750 metric calls \citep{agrawal2026gepa}, with three optimization cases sampled per iteration. 
GEPA optimization was performed on a single NVIDIA DGX Spark and required approximately 4--5 days. The controller, evaluator, and reflector were executed sequentially during optimization to remain within a single-machine setup.

\paragraph{External Evaluation.} We further assess ERIA on two independent EEG benchmarks, neither of which contributes examples to GEPA optimization or policy selection.

\underline{CerebraGloss-Bench}: This benchmark contains 90 10-second, 19-channel EEG segments designed for fine-grained clinical interpretation \citep{gu2026cerebragloss}. Although the benchmark was introduced for vision-language models operating on rendered EEG waveforms, ERIA instead analyzes the raw EEG signals to test whether its tool-based analysis policy transfers to the clinical interpretation setting. We report results on the benchmark's three language-based tasks: multiple-choice clinical interpretation, free-text EEG description, and conversational question answering, using accuracy, ROUGE-1, and the benchmark's LLM-based conversational score, respectively. The image-space waveform-localization task is excluded because ERIA does not produce image-space bounding boxes.

\underline{BrainBench}: This benchmark evaluates instruction-conditioned EEG analysis across heterogeneous analytical workflows and is organized into four subsets: Foundational Analysis, Sleep Assessment, Neurocognitive Assessment, and Physiological Integration \citep{zhou2026brainbench}. Here we evaluate ERIA on \emph{Foundational Analysis}, comprising 40 tasks and 950 instances spanning recording inspection, preprocessing, feature extraction, spectral and spatiotemporal analysis, connectivity, and evidence-grounded interpretation. This subset most directly matches ERIA's intended scope of general and clinical EEG analysis, whereas the remaining subsets target capabilities that extend beyond ERIA's intended scope. We follow the official BrainBench evaluation protocol and report its difficulty-weighted subset score on a 0--100 scale. External evaluation of the frozen ERIA agent was performed on a single NVIDIA RTX 4090. The larger evaluator and reflector are used only during offline policy optimization and are not required for deployment-time execution of ERIA.

\section{Results and Discussion}
\label{sec:results}

\paragraph{Clinical Report Quality.}

Table~\ref{tab:clinical_behavior} summarizes both the clinical evaluation and the changes in ERIA's execution behavior after GEPA policy optimization. On the validation partition, the aggregate clinical score increases from 0.5099 to 0.6277, corresponding to an absolute gain of 0.1178 and a 23.1\% relative increase. More importantly, the improvement persists on the protected test partition; its test score rises from 0.5052 to 0.6024, an absolute improvement of 0.0972 and a 19.2\% relative increase over the raw policy. The similar improvement on validation and protected test cases indicates that the selected policy transfers beyond the examples used for policy selection.

\begin{table}[t]
\caption{Clinical performance and behavioral changes after GEPA optimization. Panel A reports the aggregate physician-aligned clinical score. In Panel B, tool-call counts, controller iterations, and calls per iteration are averages per case across the 50 validation cases; percentages denote the fraction of cases or retrieval queries satisfying the indicated criterion.}
\label{tab:clinical_behavior}
\centering
\small
\setlength{\tabcolsep}{5pt}
\begin{tabularx}{\linewidth}{@{}Xrr@{}}
\toprule
\textbf{Measure} & \textbf{Raw} & \textbf{GEPA} \\
\midrule
\multicolumn{3}{@{}l}{\textit{A. Clinical report quality}} \\
Validation aggregate score & 0.5099 & 0.6277 \\
Protected test aggregate score & 0.5052 & 0.6024 \\
\midrule
\multicolumn{3}{@{}l}{\textit{B. Validation-set execution behavior}} \\
\texttt{sleep\_wake\_staging} calls & 3.42 & 7.12 \\
Cases sampling early, middle, and late recording portions & 40\% & 98\% \\
\texttt{assess\_signal\_quality} calls & 2.18 & 5.62 \\
\texttt{hemispheric\_symmetry} calls & 1.42 & 2.30 \\
\texttt{fine\_event\_classifier} calls & 4.24 & 5.68 \\
\texttt{ied\_candidate\_screen} calls & 1.16 & 1.70 \\
\texttt{retrieve\_eeg\_knowledge} calls & 1.92 & 2.34 \\
State-related retrieval queries & 46.9\% & 63.2\% \\
Total tool calls & 33.72 & 46.78 \\
Controller iterations & 10.58 & 9.34 \\
Tool calls per iteration & 3.20 & 5.19 \\
\bottomrule
\end{tabularx}
\end{table}

The trajectory comparison shows that the improvement is accompanied by a substantial change in how ERIA gathers evidence. The clearest difference is more systematic coverage of behavioral state. The optimized policy explicitly encourages examination of early, middle, and late portions of the recording and greater attention to state transitions. As a result, the proportion of validation cases in which sleep/wake evidence is sampled from all three portions increases from 40\% to 98\%, while use of \texttt{sleep\_wake\_staging} more than doubles.

Technical-quality assessment also becomes considerably more active. Instead of allowing a recording-level quality concern to dominate the interpretation, the optimized policy encourages localization of poor-quality channels or intervals and continued analysis of the usable signal. This behavior is reflected in the increase from 2.18 to 5.62 \texttt{assess\_signal\_quality} calls per case and in the policy's explicit distinction between technical limitation and clinical abnormality.

GEPA also increases corroboration of potentially abnormal findings. The selected instructions discourage treating low interhemispheric correlation as sufficient evidence of focal abnormality and direct the controller to investigate disagreement between coarse and fine event classifiers. Correspondingly, use of \texttt{hemispheric\_symmetry}, \texttt{fine\_event\_classifier}, and \texttt{ied\_candidate\_screen} all increases. The optimized policy further distinguishes physiological slowing during drowsiness or sleep from pathological slowing in a clearly awake background, making the final Normal/Abnormal decision more dependent on clinically supported findings rather than isolated detector outputs.

Knowledge retrieval becomes more targeted as well. State-related queries account for 63.2\% of retrieval calls after optimization, compared with 46.9\% under the raw policy, with the selected policy placing greater emphasis on state-dependent background characteristics, normative interpretation, and terminology relevant to epileptiform or ictal findings.

Overall, GEPA changes the allocation of analysis rather than simply increasing the length of the reasoning process. ERIA performs 46.78 tool calls per case after optimization compared with 33.72 under the raw policy, yet the mean number of controller iterations decreases from 10.58 to 9.34. Tool calls per iteration therefore rise from 3.20 to 5.19. The optimized controller gathers broader and more corroborative evidence within fewer reasoning rounds.

\paragraph{External EEG Transfer.} We next examine whether a policy optimized exclusively from whole-recording clinical reports remains useful on independent EEG tasks. Table~\ref{tab:external_results} summarizes performance on CerebraGloss-Bench and BrainBench Foundational Analysis.

\begin{table}[t]
\caption{External EEG evaluation. The reference column reports CerebraGloss-3B for CerebraGloss-Bench and the strongest reported Foundational Analysis score for BrainBench.}
\label{tab:external_results}
\centering
\small
\setlength{\tabcolsep}{4pt}
\begin{tabularx}{\linewidth}{@{}Xlrrr@{}}
\toprule
\textbf{Benchmark} & \textbf{Metric} & \textbf{Raw} &
\textbf{GEPA} & \textbf{Reference} \\
\midrule
CerebraGloss-Bench & MCQ accuracy (\%) & 71.11 & 71.11 & 80.00 \\
CerebraGloss-Bench & Description ROUGE-1 & 30.70 & 30.17 & 44.19 \\
CerebraGloss-Bench & Conversational QA & 3.60 & 3.87 & 4.13 \\
BrainBench FA & Weighted score & 85.86 & 86.89 & 84.97 \\
\bottomrule
\end{tabularx}
\end{table}

\underline{CerebraGloss-Bench}: The optimized policy preserves MCQ accuracy at 71.11\% while improving conversational QA from 3.60 to 3.87. Description ROUGE-1 decreases modestly from 30.70 to 30.17. One raw-policy case failed to produce a valid output; following the CerebraGloss-Bench evaluation convention, this case was excluded from score aggregation rather than assigned a score of zero.

The benchmark also represents a substantially different analysis setting from ERIA's optimization task. CerebraGloss-3B is explicitly post-trained for fine-grained EEG interpretation from rendered waveform images, whereas ERIA is optimized for whole-recording clinical reporting and accesses the underlying signal through specialist analytical tools rather than visual waveform input. We therefore use CerebraGloss-Bench as a transfer test rather than as a comparison between equivalent model architectures. For context, CerebraGloss-3B was the strongest model across these tasks in the original benchmark, with MCQ, ROUGE-1, and QA scores of 80.00, 44.19, and 4.76, respectively, while GPT-5 was among the strongest general-purpose baselines with corresponding scores of 70.00, 37.07, and 4.58 \citep{gu2026cerebragloss}.

The conversational QA score requires an additional qualification because it depends on an LLM judge. The released CerebraGloss evaluator uses GPT-5 to score responses against the expert reference. Re-evaluating the released CerebraGloss-3B predictions with the current version of that protocol produces a score of 4.13 rather than the originally reported 4.76. Because the released evaluator calls an unpinned model endpoint, historical and current QA scores are not guaranteed to be directly comparable. We therefore use the reproduced 4.13 score when comparing against ERIA's current results and retain 4.76 only as the published historical value.

\underline{BrainBench Foundational Analysis}: Here, the raw ERIA policy obtains a difficulty-weighted score of 85.86, while the GEPA-selected policy reaches 86.89, an absolute improvement of 1.03 points. The strongest Foundational Analysis score reported in the original BrainBench evaluation is 84.97 under CodeAct, while the strongest BrainAgent result is 81.47 \citep{zhou2026brainbench}. Both ERIA configurations exceed the highest score reported in that evaluation.

This comparison is most appropriately interpreted at the system level. BrainBench performance depends on the language model together with its execution strategy and analytical capabilities, rather than on the controller in isolation. ERIA similarly uses a capability-oriented benchmark interface while retaining its single-controller architecture. The absolute score therefore reflects the complete controller--tool system, whereas the difference between the raw and GEPA policies more specifically measures the effect of the learned analysis strategy.

\paragraph{Discussion and Limitations.}

The protected-test improvement, together with the trajectory analysis, indicates that optimizing the natural-language policy materially changes how ERIA allocates analytical effort rather than merely altering report phrasing. In particular, the selected policy favors broader temporal coverage, state-aware interpretation, and corroboration of uncertain evidence. The external evaluations further suggest that these improvements largely transfer beyond the clinical-reporting objective: GEPA improves conversational QA on CerebraGloss-Bench and Foundational Analysis on BrainBench, preserves MCQ accuracy, and incurs only a modest decrease in description ROUGE-1. Some limitations should be considered when interpreting these results. First, the GEPA-selected policy introduced one instruction that was not supported by ERIA's available specialist capabilities: it encouraged inspection of sleep spindles, K-complexes, and vertex waves despite the absence of a dedicated detector for these graphoelements. We retain the selected policy unchanged for the reported evaluations rather than manually correcting this instruction after observing it. This case illustrates that reflective policy optimization can improve analysis behavior while still introducing instructions that are inconsistent with the current toolset. Future implementations could validate proposed policy instructions against machine-readable tool contracts before deployment.

Second, the physician-aligned score measures agreement with a clinical reference report rather than independently establishing waveform-level ground truth. Reference reports may omit findings, and two clinically acceptable interpretations can differ in wording, emphasis, or level of detail. The evaluator accommodates unavailable domains and scores reports at the level of clinical concepts, but direct review by EEG specialists remains necessary to establish real-world clinical utility.

Finally, the clinical evaluation is retrospective and based on a single private collection of 200 matched EEG--report studies, including 100 protected test cases. Evaluation across additional institutions, acquisition environments, patient populations, and reporting styles will be necessary to establish the robustness of the learned policy. The external benchmarks broaden the evaluation across datasets and tasks, but they do not replace prospective multi-site clinical assessment.

\section{Conclusion}

We presented ERIA, a tool-using framework for clinical EEG reporting in which a local language-model controller coordinates specialized EEG analyses and synthesizes their outputs into a final interpretation. Optimizing only the controller's natural-language policy with GEPA increased the protected-test physician-aligned score from 0.5052 to 0.6024 without retraining the controller or specialist models. Trajectory analysis indicates that this improvement is accompanied by broader recording coverage and greater corroboration of uncertain findings, rather than a change in report phrasing alone. The optimized policy also transfers beyond the clinical-report objective, improving conversational QA on CerebraGloss-Bench and increasing BrainBench Foundational Analysis score. These findings support policy-level optimization as a practical way to improve how grounded biomedical agents deploy existing analytical capabilities without increasing controller scale or retraining their underlying models. Future work should extend the evaluation to prospective, multi-institutional settings and include direct assessment by EEG specialists rather than relying solely on automated report-level evaluation.

\subsection*{AI use statement}
Generative AI tools were used to aid and polish writing, support literature retrieval and discovery, and assist with research ideation and execution, including feedback on research methodology and experimental design. All AI-assisted content was reviewed and verified by the authors, who take full responsibility for the final manuscript, analyses, and reported results.

\subsection*{Ethics statement}
Our institutional review board approved the study in which we utilized EEG recordings and respective clinical reports of patients who opted in for retrospective data use for research.

\subsection*{Reproducibility statement}
We support reproducibility by reporting the ERIA architecture, specialist-tool provenance, clinical evaluation protocol, GEPA optimization procedure, model configurations, data partitions, and external benchmark protocols in the main text. The appendix provides additional implementation and preprocessing details, specialist-model training and evaluation procedures, GEPA implementation details, and the BrainBench evaluation interface. Source code, including prompt configurations and evaluation scripts, will be made publicly available upon acceptance. Public datasets and benchmark resources used by the specialist models and external evaluations are identified with their corresponding sources; the private clinical EEG--report collection used for the internal evaluation is not distributed with the submission.

\subsubsection*{Acknowledgments}
This work was supported by an institutional seed grant and a PhD student fellowship.

\bibliography{iclr2027_conference}
\bibliographystyle{iclr2027_conference}

\appendix
\section{Implementation and Reproducibility Details}
\label{app:implementation}

\paragraph{Retrieval Configuration.}
The \texttt{retrieve\_eeg\_knowledge} tool performs local semantic retrieval over a curated EEG knowledge base using \texttt{all-MiniLM-L6-v2} sentence embeddings \citep{reimers2019sentencebert}. Documents are divided at semantic boundaries into chunks with a target size of approximately 500 characters. Embeddings are normalized and indexed using inner-product similarity, and the three highest-scoring passages are returned for each query. The knowledge base covers EEG reporting and scope, background activity and behavioral state, interictal epileptiform activity, ictal and status terminology, normal variants and artifacts, and sleep/state interpretation.

\paragraph{Signal Quality and Preprocessing.}
ERIA exposes signal-quality information to the controller so that technical limitations can be incorporated into subsequent interpretation. The \texttt{assess\_signal\_quality} analysis checks for non-finite values, low-variance channels, and severe amplitude clipping. Any infinite value causes the corresponding analysis to be rejected. A channel is also rejected when more than 5\% of its samples are missing; smaller fractions of missing samples are imputed using the finite channel mean and the intervention is reported to the controller. Channels with standard deviation below $0.1\,\mu$V are treated as flat and rejected, whereas channels below $1.0\,\mu$V are flagged as near-flat. Samples with absolute amplitude at or above $800\,\mu$V are considered clipped. Clipping affecting at least 5\% of samples generates a quality warning, whereas clipping affecting at least 20\% causes the analysis to be rejected.

\paragraph{Additional Agent Capabilities.}
Beyond the EEG-analysis tools, ERIA supports patient-specific notes that can be retrieved in later interactions, global rules or preferences derived from recurring clinician feedback, and targeted clarification requests when additional information is required. These capabilities are disabled during GEPA optimization and all benchmark evaluations so that reported performance depends only on the current case, EEG-analysis tools, and local knowledge retrieval.

\section{Specialist Model Training and Evaluation}
\label{app:specialists}

The learned tools were developed using public EEG datasets. The reported metrics characterize the specialist models on their respective development datasets; these models serve as intermediate evidence sources within ERIA rather than as independently evaluated clinical diagnostic systems.

\paragraph{Background Abnormality.}
The \texttt{background\_abnormality} model uses ShallowFBCSPNet \citep{schirrmeister2017deep} trained on TUAB v3.0.1 \citep{lopez2015abnormal,obeid2016tuh}. The training partition contains 2,717 approximately balanced recordings, and the predefined evaluation partition contains 276 recordings. EEG is represented using 19 referential channels and non-overlapping 30-second windows. Signals are filtered from 0.5--70 Hz, notch filtered at 60 Hz, resampled to 250 Hz, and standardized per channel and window. The selected model achieves an evaluation accuracy of 84.42\% and F1 score of 80.54\% for the abnormal class.

\paragraph{Coarse Activity Screening.}
The \texttt{coarse\_activity\_screen} uses a frozen pretrained SPaRCNet encoder \citep{jing2023sparcnet} with a multinomial logistic-regression head for background, slowing, and seizure-like activity. The model operates on 10-second windows in a 16-channel double-banana bipolar montage at 200 Hz. Training combines examples from TUSL, TUSZ, and TUAR \citep{vonweltin2017slowing,shah2018tusz,hamid2020tuar}. Evaluation uses five-fold patient-group cross-validation on TUSL, with TUSZ and TUAR examples included only in the training portion of each fold. Mean accuracy is $0.675\pm0.094$ and mean macro-F1 is $0.664\pm0.101$. Classwise F1 scores are 0.627 for background, 0.632 for slowing, and 0.800 for seizure-like activity.

\paragraph{Fine Event Classification.}
The \texttt{fine\_event\_classifier} is trained on TUSZ v2.0.5 and TUAR \citep{shah2018tusz,hamid2020tuar} to distinguish background, artifact, and seizure-like activity. EEG is represented as 2.5-second windows using a 22-channel TCP bipolar montage at 250 Hz. Each window is described by handcrafted amplitude, statistical, Hjorth, temporal, spectral, and inter-channel features and classified using XGBoost \citep{chen2016xgboost}. Training and evaluation partitions are patient-disjoint. The model achieves an evaluation accuracy of 0.710 and macro-F1 of 0.691, with classwise F1 scores of 0.742 for background, 0.571 for artifact, and 0.760 for seizure-like activity.

\paragraph{Sleep--Wake Staging.}
The \texttt{sleep\_wake\_staging} model uses SleepStagerChambon2018 \citep{chambon2018sleep} trained on Sleep-EDF Expanded \citep{kemp2000sleepedf} and EEGMMIDB \citep{schalk2009eegmmidb}. It predicts six states: eyes-open awake, eyes-closed awake, drowsiness, light sleep, deep sleep, and REM sleep. Inputs consist of Fpz--Cz and Pz--Oz derivations sampled in 30-second windows and resampled to 100 Hz. A subject-grouped 80/20 split is used with no subject overlap between training and evaluation. The selected model achieves an evaluation accuracy of 0.763 and macro-F1 of 0.764. Classwise F1 scores are 0.854 for eyes-open awake, 0.820 for eyes-closed awake, 0.583 for drowsiness, 0.833 for light sleep, 0.762 for deep sleep, and 0.732 for REM sleep. When Oz is unavailable at inference time, the mean of O1 and O2 is used as an occipital substitute for constructing Pz--Oz, with the substitution reported to the controller.

\paragraph{IED Candidate Screening.}
The \texttt{ied\_candidate\_screen} uses the released SpikeNet2 checkpoint \citep{li2025spikenet2} without additional ERIA-specific training or fine-tuning. EEG is represented using 19 referential electrodes, resampled to 128 Hz, and transformed according to the released SpikeNet2 preprocessing procedure into 37 derived channels. Screening uses overlapping 1-second windows with a 0.25-second stride. ERIA retains the continuous SpikeNet2 probability for interpretation; a threshold of 0.50 is used only to identify and merge candidate intervals.

\section{GEPA Implementation Details}
\label{app:gepa_details}

Experiments use GEPA v0.1.4 \citep{agrawal2026gepa}. Beyond the optimization settings reported in the main text, reflector decoding uses temperature 1.0, top-$p$ 0.95, top-$k$ 20, a presence penalty of 1.5, and a maximum generation length of 32,768 tokens. The clinical evaluator uses schema-constrained JSON output with a maximum generation length of 12,288 tokens.

The rendered input to each reflection call is limited to 90,000 tokens. The target reflection minibatch contains three complete execution trajectories. When all three trajectories cannot fit within this budget, the reflection context is reduced to two and, if necessary, one complete trajectory rather than truncating individual trajectories. This adaptation affects only the context supplied to the reflector; candidate-policy evaluation continues to use the originally sampled GEPA minibatch. In the reported optimization run, this fallback was invoked once, reducing the reflection context from three trajectories to two.

\section{External Benchmark Adaptations}

\paragraph{BrainBench.}
Following the BrainBench adaptation protocol \citep{zhou2026brainbench}, we augmented ERIA with additional general-purpose deterministic EEG-analysis capabilities required for Foundational Analysis, including extended spectral measures, signal statistics, connectivity analysis, filtering and signal-transformation analysis, and generation of analysis artifacts. These capabilities were not required for ERIA's primary whole-recording report-generation task and are not specialized to individual BrainBench tasks.

\paragraph{CerebraGloss-Bench.}
CerebraGloss-Bench provides 10-second EEG segments rather than whole recordings \citep{gu2026cerebragloss}. Consequently, several ERIA tools designed for longer temporal contexts, including those requiring 30-second windows or broader recording-level coverage, were not applicable during this evaluation. ERIA therefore operated using only the subset of its tools compatible with the available 10-second segments.

\end{document}